\documentclass{article}
\usepackage{spconf,amsmath,epsfig}
\usepackage{amsfonts}
\usepackage{amssymb}
\usepackage[font=small,labelfont=bf]{caption}
\usepackage{breqn}
\title{Visual Question Answering in Remote Sensing with Cross-Attention and Multimodal Information Bottleneck}
\name{Jayesh Songara$^1$, Shivam Pande$^2$, Shabnam Choudhury$^2$, Biplab Banerjee$^2$, Rajbabu Velmurugan$^1$ \thanks{Corresponding author: shivam\_pande@iitb.ac.in} \thanks{$^1$Department of Electrical Engineering, IIT Bombay, India} \thanks{$^2$Centre of Studies in Resources Engineering, IIT Bombay India}}

\address{Indian Institute of Technology Bombay, India}
\begin{document}
%\ninept
%
\maketitle
\begin{abstract}
In this research, we deal with the problem of visual question answering (VQA) in remote sensing. While remotely sensed images contain information significant for the task of identification and object detection, they pose a great challenge in their processing because of high dimensionality, volume and redundancy. Furthermore, processing image information jointly with language features adds additional constraints, such as mapping the corresponding image and language features. To handle this problem, we propose a cross attention based approach combined with information maximization. The CNN-LSTM based cross-attention highlights the information in the image and language modalities and establishes a connection between the two, while information maximization learns a low dimensional bottleneck layer, that has all the relevant information required to carry out the VQA task. We evaluate our method on two VQA remote sensing datasets of different resolutions. For the high resolution dataset, we achieve an overall accuracy of 79.11\% and 73.87\% for the two test sets while for the low resolution dataset, we achieve an overall accuracy of 85.98\%. 
\end{abstract}
\section{Introduction}
\label{sec:intro}

With the advancement in sensing technology, significant visual information is being acquired and processed in the remote sensing domain. The information, when efficiently processed gives reliable results for tasks like image classification, object detection \cite{ma2019deep}. These problems seem to be more static in nature and have been handled very well using techniques from deep learning. However, these problems are more task-specific in nature i.e. a problem defined for the task of road detection might not be suitable for other tasks. This issue can be circumvented by framing the problem in a more practical scenario that would consider different aspects within the same problem-set. This is where visual question answering (VQA) comes into picture \cite{lobry2020rsvqa}. In this scenario, the information can be obtained from the images by making queries on the images to get relevant answers. The problem poses a challenge for machines because of their limited ability to decode the semantic and contextual meaning in the query. Prior research in computer vision has addressed these issues with both machine and deep learning techniques \cite{manmadhan2020visual}.  

VQA was addressed in remote sensing much later due to large amount of data, domain differences for multimodal data and relatively lower image resolution. \cite{lobry2020rsvqa}. Fig. \ref{fig:sample2} shows an example of visual queries addressed in remote sensing images. \cite{lobry2020rsvqa} uses convolutional neural networks (CNN) and recurrent neural networks (RNN) for visual and semantic feature extraction, and fuse them to get a joint representation. However, using such approaches has a drawback that it takes both significant and irrelevant features into account, leading to subpar embedding, resulting in suboptimal performance. This kind of problem can be addressed using the idea of attention and information based learning \cite{hjelm2018learning}. In the former, the model learns the attention mask (a tensor) that explicitly highlights the relevant features in the data. This technique is further improved by using cross-attention \cite{pande2021adaptive}, where the masks generated from one modality filters out irrelevant features of other modality. In the latter, we maximize the mutual information between the two modalities to get more informative embedding. 

\begin{figure*}[ht!]
  \centering
  \centerline{\includegraphics[width=14cm]{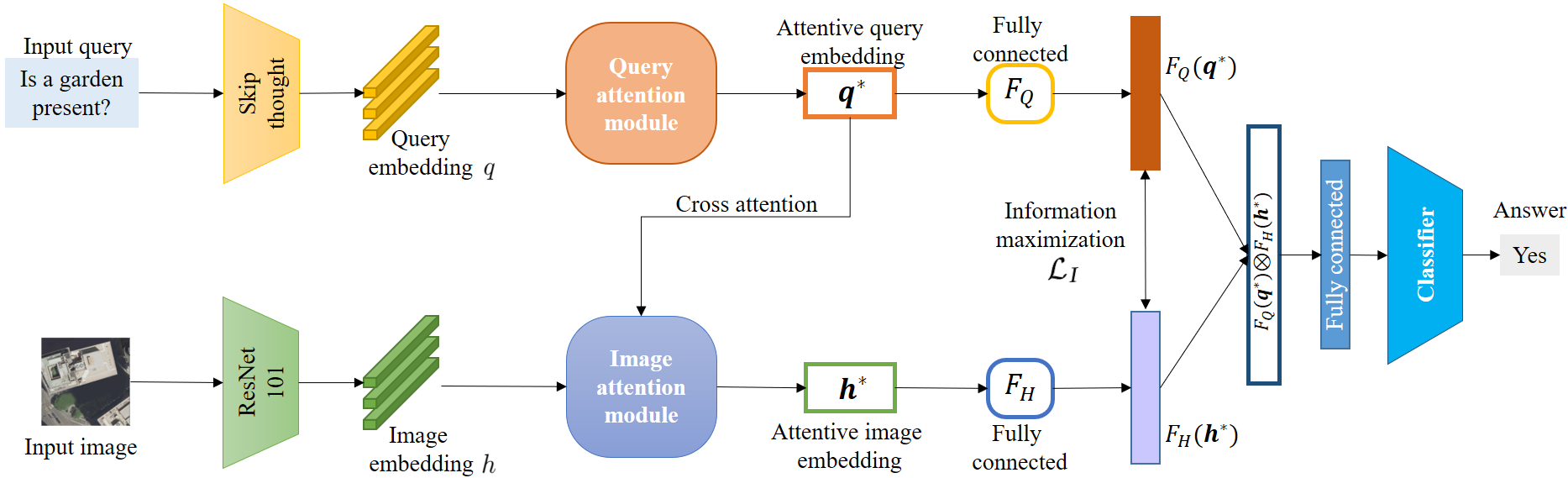}}
\caption{Schematic of the proposed VQA model. The query embedding is sent to the query attention module to get attentive features. The image embedding is sent to the image attention module, and using cross connection from the attentive query embedding, attentive image embedding is obtained. The two embedding are passed through FC layers, and the resulting outputs are subjected to information maximization loss.}\medskip
%\caption{Schematic of proposed VQA model. The image and the query are respectively sent to pretrained ResNet 101 and Skip thought models and converted into embeddings. The query embedding is sent to the query attention module to get attentive features. The image embedding is sent to the image attention module, and using cross connection from the attentive query embedding, attentive image embedding is obtained. The two embedding are passed through FC layers, and the resulting outputs are subjected to information maximization loss. The outputs from the FC layers are multiplied and sent for classification.}\medskip
  \vspace{-0.5cm}
\label{fig:Diagram}
\end{figure*}

In our method, we approach mutual information maximization technique in a multimodal setting, where the learned representation tries to capture all the relevant information and simultaneously discard the superfluous information with respect to the query response. The information for both the visual and language modalities are encoded in a latent vector of relatively lower dimensions using the bottleneck layer \cite{federici2020learning}. This enables us to learn relevant features, leading to more robust predictions. We adapt the idea of cross attention for VQA as presented in \cite{li2020visual} coupled with multimodal information maximization through a bottleneck layer. We evaluate our method on two most recent VQA datasets in remote sensing and achieve state of the art results. Our contributions are:

\begin{itemize}
    \item We introduce multimodal information bottleneck for VQA in remote sensing. The bottleneck layer captures the relevant information from visual and language modalities while rejecting the redundant information. 
    \item We introduce the idea of cross-attention for VQA in remote sensing to highlight relevant features across modalities for robust learning. 
\end{itemize}

\section{Methodology}

For VQA, we consider a dataset defined as a pair of images and corresponding query given as $\mathcal{X} = \{\textit{\textbf{x}}^i_I, \textit{\textbf{x}}^i_Q\}_{i=1}^n$ and the corresponding answers $\mathcal{Y} = \{y^i\}_{i=1}^n$. Here, $\textit{\textbf{x}}^{{i}}_I \in \mathbb{R}^{M \times N \times B}$ represents the image ($M$, $N$ and $B$ respectively being the number of rows, columns and bands in the image) and $\textit{\textbf{x}}^i_Q = \{x^i_{q_1}, x^i_{q_2},...,x^i_{q_K}\}$ represents the query on the image, with {$K$ being the number of words in the query.} Here, $n$ is the total number of samples, while $i$ is the index of the sample. The framework is divided into cross-attention based feature extractor and mutual learning assisted classifier that are explained {in the subsequent sections.}

\subsection{Feature extraction}
\label{sec:FE}

The image and query samples are converted into embeddings using CNN and LSTM based feature extractors ($G$ and $H$ respectively). Initially, all the image samples are converted into an embedding of dimensions {$(T \times 2048)$} {as presented in} 
\begin{equation}
{\textit{\textbf{h}}} = G(\textit{\textbf{x}}_I),
\label{equation:ResNet}
\end{equation}
where, $\textit{\textbf{h}}$ is the set of extracted objects and are represented as $\textit{\textbf{h}} \in \{h_1, h_2,...,$
$h_T\}$. Here, {$T$} denotes the number of objects that are extracted from the image. Simultaneously, the queries are encoded into a tensor with dimensions $(K \times 2400$) as shown in
\begin{equation}
{\textit{\textbf{q}}} = H(\textit{\textbf{x}}_Q)
\label{equation:skip}
\end{equation}

Here, $H$ is the skip thoughts model and $\textit{\textbf{q}}$ is embedding for query $\textit{\textbf{x}}^i_Q$. The resulting features are then combined using a gating mechanism across the text and image modalities. They are discussed in the following sections.

\subsubsection{Query attention features} 

\noindent To harness the characteristic of attention mechanism, we try to focus on the important keywords in the question using attention weights learnt from question representations, as in 
{\begin{equation}
{a}^{q}_{k} = J_{a^q}ReLU(w_{a^q}q_k),
\label{equation:qJ}
\end{equation}}
{\begin{equation}
\alpha^{q} = softmax({\textit{\textbf{a}}}^{q}),
\label{equation:q_alpha}
\end{equation}}

\noindent where, $k$ refers to the index of the word, $J_{a^q} \in \mathbb{R}^{d_{ff} \times 1}$, $w_{a^q} \in \mathbb{R}^{d_q \times d_{ff}}$, $d_{ff}$ and $d_q$ are dimension of layers and length of each word in embedding vector respectively, $q_k$ refers to the query embeddings, that are created using the RNN based model and $\alpha$ is the attention weight. Finally, the question representations are weighted to obtain the attention enhanced representations given in
\begin{equation}
\textit{\textbf{q}}^* = \sum_{k=1}^{K} \alpha_{k}^{q}q_{k},
\label{equation:q_star}
\end{equation}
where $\textit{\textbf{q}}^*$ is the attention enhanced query embedding.

\subsubsection{Image attention features} 

%\noindent\textbf{Image attention features}

\noindent This is a two step process. Initially, we employ attention on images and get the image embeddings
\begin{equation}
h'_t = (w_hh_t) \otimes (w_{q^*}\textit{\textbf{q}}^*)
\label{equation:imgatt1}
\end{equation}
The symbols {$w_h\in\mathbb{R}${$^{d_p}$}$^{\times d_h}$} and {$w_{q^*}\in\mathbb{R}${$^{d_p}$}$^{\times d_q}$} represent weights of object and query embedding, respectively, while $\otimes$ is the Hadamard product, $t$ is the object index {and $d_p$ is the common dimension of the projection of image and query embedding to enable their combination.}  

The fused representation $h_{t}'$ is used to predict the importance of $t^{th}$ object for the corresponding query as shown in 

\begin{equation}
a^{h}_t = J_{a^h}ReLU(w_{a^h}h_{t}'),
\label{equation:imgU}
\end{equation}

\begin{equation}
\alpha^{h} = softmax({\textit{\textbf{a}}}^h),
\label{equation:img_alp}
\end{equation}

\noindent where, $w_{a^h} \in \mathbb{R}^{d_{ff} \times}${$^{{d_{p}}}$} and $J_{a^h} \in \mathbb{R}^{d_{ff} \times 1}$ and $\alpha^h$ are the image attention weights. The attention enhanced image object embedding is shown below as
\begin{equation}
\textit{\textbf{h}}^* = \sum_{t=1}^{T} \alpha_{t}^{h}h_{t}
\label{equation:img_star}
\end{equation}

\subsubsection{Feature fusion and classification} 

\noindent The output size of $\textit{\textbf{q}}^*$ and $\textit{\textbf{h}}^*$ are $1 \times 2048$ and $1 \times 2400$, respectively. Hence, they are passed through two fully connected layers to get size of $1 \times 1200$ each and the outputs are combined using Hadamard product. This can be represented as {$F_Q(\textit{\textbf{q}}^*) \otimes F_H(\textit{\textbf{h}}^*)$}, where $F_Q$ and $F_H$ are fully connected layers for query and image embeddings respectively. The {resulting} vector is passed through a multilayer perceptron model for classification, where the class corresponds to the answer to the posed query. The problem has been posed as the classification one (akin to \cite{lobry2020rsvqa}) because the answers in the groundtruth belong to the predefined categories {(such as \textit{presence}, \textit{count}, \textit{comparison}, \textit{area}, \textit{rural/urban})}, and hence, categorical cross entropy loss is employed as    
\begin{equation}
\mathcal{L}_{CE} = -\sum_{i=1}^{N_Q} y^i\log{\hat{y}^i},
\label{equation:ce}
\end{equation}
where $N_Q$ is the number of query samples and $\hat{y}^i$ is the predicted label for $i^{th}$ sample.

\subsection{Training by information maximization}
\label{sec:mmbn}

This technique is used to maximize the interaction and information between the image and the query. This helps the model to focus on the objects represented in the query in the context of the image.  This is achieved by jointly projecting the features of the image and query embedding into a low dimensional common feature space (called \textit{bottleneck}, due to the transformation from higher to much lower dimensions \cite{federici2020learning}). Let $z_{\textit{\textbf{q}}^*}$ and $z_{\textit{\textbf{h}}^*}$ represent the corresponding minimal latent representation for $F_Q(\textit{\textbf{q}}^*)$ and $F_H(\textit{\textbf{h}}^*)$, respectively. Losses $\mathcal{L}_1$ for $F_H(\textit{\textbf{h}}^*)$ and $\mathcal{L}_2$ for $F_Q(\textit{\textbf{q}}^*)$ are presented as  
\begin{equation}
\mathcal{L}_1(\phi, \lambda_1) = I_{\phi}(z_{\textit{\textbf{q}}^*}; F_Q(\textit{\textbf{q}}^*)|F_H(\textit{\textbf{h}}^*)) - \lambda_1I_{\phi}(F_H(\textit{\textbf{h}}^*);z_{\textit{\textbf{q}}^*}),
\label{equation:rep1}
\end{equation}
\begin{equation}
\mathcal{L}_2(\psi, \lambda_2) = I_{\psi}(z_{\textit{\textbf{h}}^*}; F_H(\textit{\textbf{h}}^*)|F_Q(\textit{\textbf{q}}^*) ) - \lambda_2I_{\psi}(F_Q(\textit{\textbf{q}}^*);z_{\textit{\textbf{h}}^*}),
\label{equation:rep2}
\end{equation}
where, $\phi$ and $\psi$ are the parameters of question and image encoders built on top of latent question and image embedding, $\lambda_1$ and $\lambda_2$ are the Lagrangian parameters corresponding to the mutual information $I_\phi$ and $I_\psi$.
The joint representation for the two losses is calculated by averaging the losses $\mathcal{L}_1$ and $\mathcal{L}_2$. An upper bound is established on the loss by using symmetrised KL divergence ($D_{SKL}$). The final loss is
\begin{dmath}
\mathcal{L}_I= -I_{\phi\psi}(z_{\textit{\textbf{q}}^*}; z_{\textit{\textbf{h}}^*}) + \gamma D_{SKL}(p_{\phi}(z_{\textit{\textbf{q}}^*}|F_Q(\textit{\textbf{q}}^*))||p_{\psi}(z_{\textit{\textbf{h}}^*}|F_H(\textit{\textbf{h}}^*))),
\label{equation:info}
\end{dmath}
where, $I_{\phi\psi}$ is the average mutual information and $\gamma$ is the learnable regularization.
Our model is jointly trained using  
\begin{equation}
\mathcal{L}_{final}= \mathcal{L}_{CE}+\lambda \mathcal{L}_{I},
\label{equation:final}
\end{equation}
where, $\lambda$ is the balancing coefficient. It was observed empirically that the models work best when $\lambda$ is fixed to 1, since both are KL divergence based losses and in the context of this work, they are equally important.

\begin{figure*}[ht!]
  \centering
  \centerline{\includegraphics[width=14cm]{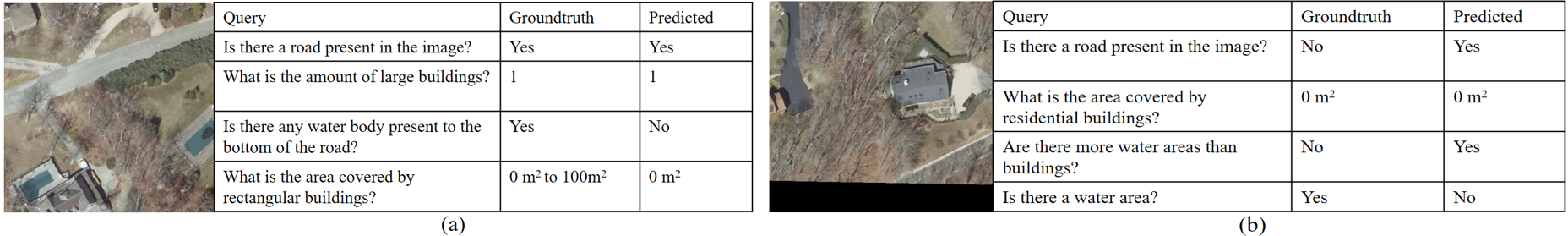}}
\caption{VQA test samples for HR dataset with groundtruth and corresponding answer with our method.}\medskip
  \vspace{-0.5cm}
\label{fig:sample2}
\end{figure*}

\section{EXPERIMENTS AND RESULTS}
\label{sec:experimets}

This section presents the datasets, experiments and a discussion over the results.

\subsection{Datasets}
\label{ssec:datasets}

We select two remote sensing datasets to evaluate our visual question answering approach. 

\noindent\textbf{High resolution (HR) dataset}: The dataset is composed of images from USGS High Resolution Orthoimagery data (samples at 15cm spatial resolution). There are 10,659 images (of spatial size 512 $\times$ 512) and 1,066,316 queries. However, because of resource constraints, only 285,000 queries were considered. The dataset is then divided into training set (61.5\%), test set 1 (31.7\%) and test set 2 (6.8\%) \cite{lobry2020rsvqa}. 

\noindent\textbf{Low resolution (LR) dataset}: The images are acquired from Sentinel 2 satellite over Netherlands with 10m spatial resolution. There are total 772 images with 256 $\times$ 256 spatial size and 77,232 queries. The ratio of division of the dataset into train and test sets is kept as 87.2\% and 12.8\% \cite{lobry2020rsvqa}.

\begin{table*}[ht]
\centering{\scriptsize
 \caption{\label{tab:HR1} Accuracy analysis on the HR dataset (\%). When both cross-attention and information bottleneck are used, we get the best performance.}
\begin{tabular}{|p{1.5cm} |p{1.1cm} |p{1.1cm} |p{1.1cm} |p{1.2cm}|p{1.1cm} |p{1.1cm} |p{1.1cm} |p{1.2cm}|}
 \hline 
 & \multicolumn{4}{c|}{Test set 1 } & \multicolumn{4}{c|}{Test set 2} \\
 \hline
Class & RSVQA \cite{lobry2020rsvqa}& CrossAtt & InfoMax & InfoMax + CrossAtt & RSVQA \cite{lobry2020rsvqa}& CrossAtt & InfoMax & InfoMax + CrossAtt\\
 \hline
Count  &52.34 & 55.56 & 55.34 & \textbf{56.22} & 48.44 & 50.39 & 49.92 & \textbf{51.22}\\
Presence  &92.54 & 93.48 & 93.16 & \textbf{94.12} & 90.04 & 88.75 & 90.46 & \textbf{91.34}\\
Comparison  &75.34 & \textbf{88.51} & 88.23 & 88.29 & 76.33 & 77.23 & \textbf{81.91} & 80.21\\
Area  & \textbf{85.32} & 82.61 & 83.96 &84.12 & 65.81 & 65.77 & \textbf{71.96}& 70.32\\
 \hline
OA  &78.23 & 78.48 & 78.50 &\textbf{79.11} & 72.45 & 70.56 & 72.95 & \textbf{73.87}\\
AA &76.38 & 80.04 & 80.16 &\textbf{80.63} & 70.15 & 70.54 & 73.27 & \textbf{73.56}\\
 \hline
\end{tabular}}
\end{table*}

\begin{table}[ht]
\centering{\scriptsize
 \caption{\label{tab:LR} Accuracy analysis on the LR dataset (\%).}
\begin{tabular}{|p{1.5cm} |p{1.3cm} |p{1.2cm} |p{1.2cm} |p{1.3cm}|}
 \hline
Class & RSVQA \cite{lobry2020rsvqa}& CrossAtt & InfoMax & InfoMax + CrossAtt\\
 \hline
Count &{65.38} & 72.21 & 70.56  & \textbf{73.12}\\
Presence&{84.78} & 91.67 & 90.12 & \textbf{92.86}\\
Comparison &{78.44} & \textbf{93.08} & 88.95 & 93.01\\
Rural/Urban &{\textbf{89.67}} & 85.00 & 85.12 & 88.12\\
 \hline
OA &{79.56} & 85.46 & 84.13 & \textbf{85.98}\\
AA &{78.34} & 85.49 & 83.68 & \textbf{86.77}\\
 \hline
\end{tabular}}
\end{table}

\subsection{Training protocols}
\label{ssec:tp}

For image and query feature extraction, ResNet-101 \cite{wu2019wider} and skip thoughts model \cite{kiros2015skip} are respectively used. We create all the models in PyTorch and train them on {6GB} NVIDIA GeForce RTX 2060 GPU. For LR dataset, the number of epochs are set to 150 while the batch size is fixed to 280. For HR datasets, the number of epochs and the batch size are kept low (35 and 70 respectively). This is due to large image size that makes their processing slow and resource intensive. The models use Adam optimizer with learning rate of $10^{-5}$. For the LR dataset, four categories of questions are addressed, namely, \textit{count}, \textit{presence}, \textit{comparison} and \textit{rural/urban}. For HR dataset, the first three categories are same as LR dataset while the fourth category is ``area". We compare our method with the most recent benchmark in remote sensing VQA (RSVQA) \cite{lobry2020rsvqa}, since it is the only prominent research available that tackles VQA in remote sensing. We also carry out ablation studies, where we remove the information maximization and cross attention, and then compare the performance using overall accuracy (OA), average accuracy (AA) and class-wise accuracy. 

\subsection{Results and Discussions}
\label{sec:rnd}

The results corresponding to the two datasets are presented in Tables \ref{tab:HR1} and \ref{tab:LR}. For both the datasets, our method shows superior performance to the RSVQA method both in overall accuracy (79.11\% and 73.87\% for test sets 1 and 2 respectively for HR dataset, and 85.98\% for LR dataset) and average accuracy (80.63\% and 73.56\% for test sets 1 and 2 respectively for HR dataset, and 86.77\% for LR dataset). It is also visible that when either of the cross attention module or information maximization is removed, the performance drops. Our approach also gives a reasonably good performance for individual classes for both the datasets. 

%\begin{figure*}[ht!]
%  \centering
%  \centerline{\includegraphics[width=14cm]{images/gc.png}}
%\caption{{VQA test samples with GradCam representations and queries. We have the original image and the GradCAM maps %created using RSVQA and ablations of our model (with only cross-attention, information maximization and both modules, %denoted as `proposed'). The answers corresponding to every map are provided below the map.}}\medskip
%  \vspace{-0.5cm}
%\label{fig:gc}
%\end{figure*}

Fig. \ref{fig:sample2} presents two samples with queries and answers from groundtruth and our proposed approach. In Fig. \ref{fig:sample2} (a), the response to the last query to calculate the area is predicted incorrect. The reason for this could be that the entire problem is treated as classification task and trained accordingly, while predicting area is more of a regression task. In Fig. \ref{fig:sample2} (b), we could see certain wrong classification results for queries 1, 3 and 4. The reason could be the presence of black area in the image that is interfering with the classification. %{In Fig. \ref{fig:sample2} (c), the response to the last two queries are incorrect, probably due to low resolution data, leading to wrong identification of buildings. The same could be the reason for the wrong predictions in the last two queries in Fig. \ref{fig:sample2} (d).}

\section{Conclusions}
\label{sec:conclusion}
We present a novel VQA approach in remote sensing. The main idea here is to discard any superfluous information and only proceed with the relevant features. To this end we propose a cross-attention based architecture governed by mutual information maximization. The former extracts the relevant embedding from the image and query domains while the latter is used to give a bounded space which only has the information necessary for answering the queries. We evaluate our method on two remote sensing datasets where our approach outperforms few of the prominent existing benchmarks. In future, we would make the problem less constrained and not consider predefined fixed set of answers.%In future, we are considering working with the VQA problem in Open World domain, where questions/answer pairs unseen during training would also be used for testing.  
%\section*{Acknowledgement}
%The authors would like to acknowledge the IITB-ISRO Grant RD/0120-ISROC00-005.

\bibliographystyle{IEEEbib}
{\footnotesize\bibliography{refs}}

\begin{thebibliography}{1}

\bibitem{ma2019deep}
Lei Ma, Yu~Liu, Xueliang Zhang, Yuanxin Ye, Gaofei Yin, and Brian~Alan Johnson,
\newblock ``Deep learning in remote sensing applications: A meta-analysis and
  review,''
\newblock {\em ISPRS journal of photogrammetry and remote sensing}, vol. 152,
  pp. 166--177, 2019.

\bibitem{lobry2020rsvqa}
Sylvain Lobry, Diego Marcos, Jesse Murray, and Devis Tuia,
\newblock ``{RSVQA}: Visual question answering for remote sensing data,''
\newblock {\em IEEE Transactions on Geoscience and Remote Sensing}, 2020.

\bibitem{manmadhan2020visual}
Sruthy Manmadhan and Binsu~C Kovoor,
\newblock ``Visual question answering: a state-of-the-art review,''
\newblock {\em Artificial Intelligence Review}, vol. 53, no. 8, pp. 5705--5745,
  2020.

\bibitem{hjelm2018learning}
R~Devon Hjelm, Alex Fedorov, Samuel Lavoie-Marchildon, Karan Grewal, Phil
  Bachman, Adam Trischler, and Yoshua Bengio,
\newblock ``Learning deep representations by mutual information estimation and
  maximization,''
\newblock {\em arXiv preprint arXiv:1808.06670}, 2018.

\bibitem{pande2021adaptive}
Shivam Pande and Biplab Banerjee,
\newblock ``Adaptive hybrid attention network for hyperspectral image
  classification,''
\newblock {\em Pattern Recognition Letters}, vol. 144, pp. 6--12, 2021.

\bibitem{federici2020learning}
Marco Federici, Anjan Dutta, Patrick Forr{\'e}, Nate Kushman, and Zeynep Akata,
\newblock ``Learning robust representations via multi-view information
  bottleneck,''
\newblock {\em arXiv preprint arXiv:2002.07017}, 2020.

\bibitem{li2020visual}
Wei Li, Jianhui Sun, Ge~Liu, Linglan Zhao, and Xiangzhong Fang,
\newblock ``Visual question answering with attention transfer and a cross-modal
  gating mechanism,''
\newblock {\em Pattern Recognition Letters}, vol. 133, pp. 334--340, 2020.

\bibitem{wu2019wider}
Zifeng Wu, Chunhua Shen, and Anton Van Den~Hengel,
\newblock ``Wider or deeper: Revisiting the {R}es{N}et model for visual
  recognition,''
\newblock {\em Pattern Recognition}, vol. 90, pp. 119--133, 2019.

\bibitem{kiros2015skip}
Ryan Kiros, Yukun Zhu, Russ~R Salakhutdinov, Richard Zemel, Raquel Urtasun,
  Antonio Torralba, and Sanja Fidler,
\newblock ``Skip-thought vectors,''
\newblock {\em Advances in neural information processing systems}, vol. 28, pp.
  3294--3302, 2015.

\end{thebibliography}

\end{document}